\documentclass[letterpaper]{article}
\usepackage[preprint]{aaai2027}
\usepackage[hyphens]{url}
\usepackage{graphicx}
\usepackage{natbib}
\usepackage{caption}
\usepackage{booktabs}
\usepackage{amsmath}
\usepackage{amssymb}
\title{Learning Gaussian Structure: Intervention-Guided Density Control for Feed-Forward Driving Reconstruction}
 \author{
      Hang Li\textsuperscript{\rm 1},
      Jiahe Li\textsuperscript{\rm 1},
      Meiying Gu\textsuperscript{\rm 1},
      Jin Zheng\textsuperscript{\rm 1,\rm 2,\rm 3}\footnote{Corresponding author: Jin Zheng (jinzheng@buaa.edu.cn)},
      Lina Yu\textsuperscript{\rm 4},
      Xiao Bai\textsuperscript{\rm 1,\rm 2}
  }
\affiliations{
    \textsuperscript{\rm 1}School of Computer Science and Engineering, Beihang University\\
    \textsuperscript{\rm 2}State Key Laboratory of Software Development Environment, Jiangxi Research Institute, Beihang University\\
    \textsuperscript{\rm 3}State Key Laboratory of Virtual Reality Technology and System, Beihang University\\
    \textsuperscript{\rm 4}AnnLab, Institute of Semiconductors, Chinese Academy of Sciences\\
    
    \{lihang13, jinzheng\}@buaa.edu.cn

}

\begin{document}
\maketitle

\begin{abstract}
Feed-forward Gaussian reconstruction has recently emerged as an efficient approach for driving scene reconstruction. However, prevailing LiDAR-based methods preserve the initial correspondence between observed points and Gaussian primitives, treating the initialized primitive set as the final representation. Unlike optimization-based 3DGS, these methods cannot accumulate gradients during training to determine how the scenes representation should be densified. Meanwhile, the shared sparse backbone only fuses observations from different timestamps implicitly, without explicitly aggregating cross-time evidence for individual primitives. In this paper, we present Learning Gaussian Structure (LGS), a framework that enhances both Gaussian structure and primitive attributes. Our key observation is that changes in local gradient responses induced by a prune or add intervention reveal whether the corresponding structural adjustment benefits reconstruction. Based on this observation, our Gaussian Densify Policy learns a Densify Map comprising Prune and Addition Scores from controlled interventions, and directly adjusts the Gaussian structure during inference. We further develop a compact Cross-Time Point Query that explicitly retrieves and aggregates neighboring features from Gaussian primitives at other timestamps for reliable attribute prediction. Extensive experiments on the Waymo and PandaSet demonstrate that LGS consistently outperforms existing methods.
\end{abstract}

\begin{figure}[t]
  \centering
   \includegraphics[width=1.\linewidth]{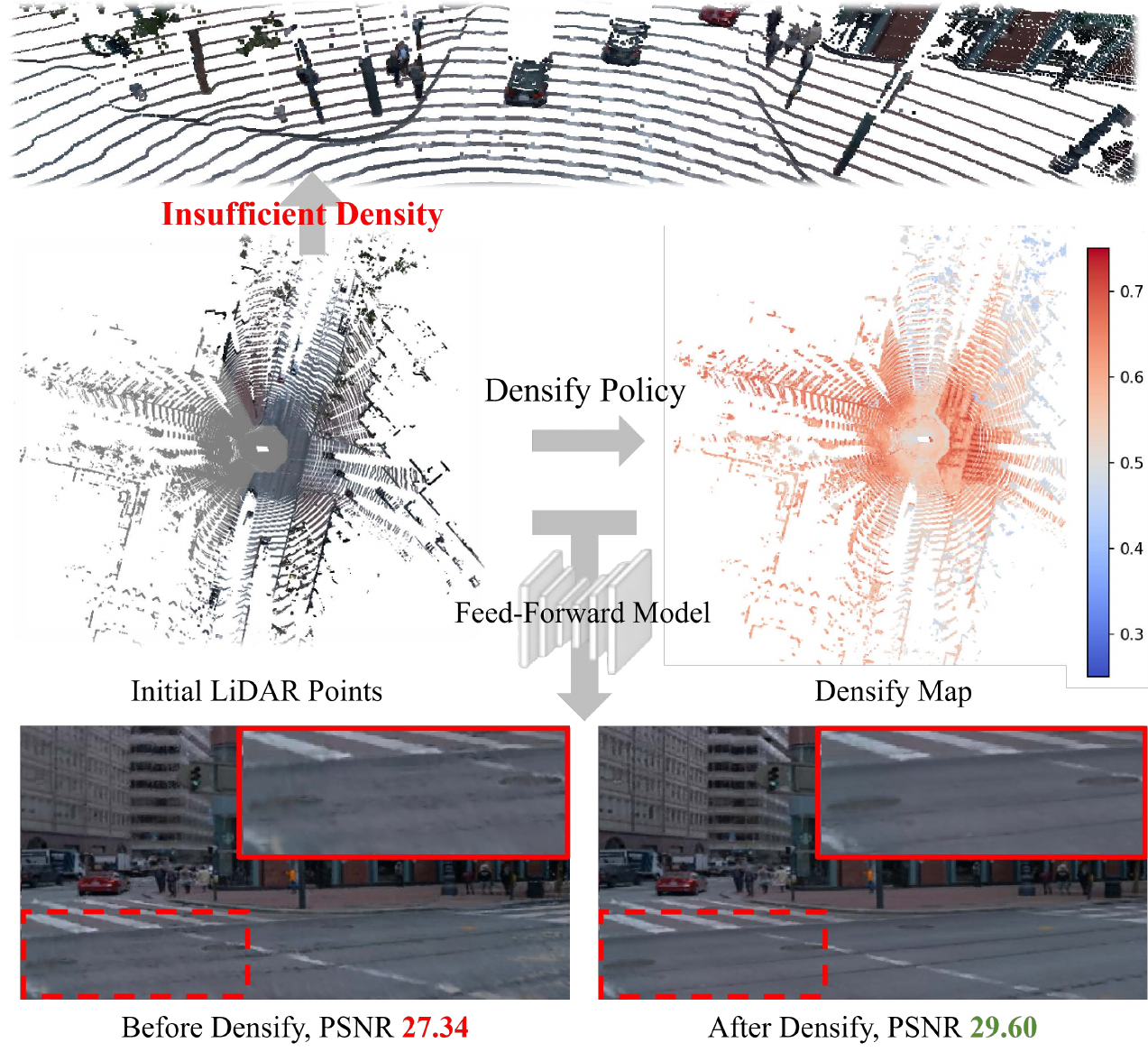}
   \caption{Learning Gaussian Structure for feed-forward driving
   reconstruction. Existing methods preserve the initial mapping from LiDAR
   points to Gaussian primitives, which can leave insufficient density in
   regions requiring additional representation capacity. The Gaussian Densify
   Policy predicts a Densify Map to restructure the Gaussian representation
   and improve reconstruction quality.}
   \vspace{-.2cm}
   \label{fig:intro}
\end{figure}

\section{Introduction}

Reconstructing dynamic driving scenes from LiDAR points captured at different timestamps and multi-camera images is fundamental to autonomous driving, simulation, and sensor synthesis. Neural radiance fields~\citep{mildenhall2020nerf} and 3D Gaussian Splatting~\citep{kerbl3dgs2023} have achieved impressive rendering quality, but most approaches require costly optimization for each scene individually~\citep{emernerf2023,desiregs2025}. Recently, feed-forward reconstruction~\citep{storm2025,unisplat2025,pointforward2026} has emerged as an efficient alternative that directly maps sensor observations to dynamic Gaussian scenes.

Prevailing feed-forward methods initialize Gaussian primitives from pixels~\citep{pixelsplat2024,mvsplat2024,splatterimage2024}, points~\citep{evolsplat2025}, voxels~\citep{unisplat2025}, or learned structures ~\citep{pointforward2026,ground4d2026}. While significant progress has been achieved, one critical limitation lies in the construction of the Gaussian representation. In LiDAR-based reconstruction, each LiDAR point initializes one Gaussian primitive, after which a 3D backbone predicts its attributes and motion. This design preserves metric geometry and enables efficient prediction. However, it also retains the initial mapping throughout inference and treats the initialized Gaussian set as the final representation. Continuous regression can update the position and other attributes of an existing primitive, but it cannot determine whether that primitive should be pruned or another primitive should be added. As illustrated in Figure~\ref{fig:intro}, the resulting sparse representation fails to recover fine scene structures and produces blurred details.

Introducing densification into feed-forward reconstruction is non-trivial. Conventional 3DGS~\citep{kerbl3dgs2023} progressively prunes and densifies primitives according to gradients accumulated during optimization. In contrast, a feed-forward model directly predicts the Gaussian representation of an unseen scene and cannot rely on such an optimization procedure. Learning Gaussian structure adjustment therefore requires both a transferable densification mechanism and a training signal that indicates whether each primitive should be pruned or added. 
To obtain such a training signal, we observe that controlled structural interventions reveal whether a Gaussian primitive benefits the local representation. Pruning or adding a primitive redistributes the rendering gradients among its spatial neighbors. An operation that reduces the local gradient response indicates that the surrounding representation can better explain the observations. These responses therefore provide direct supervision for learning structural adjustment.
A separate limitation concerns reliable Gaussian attribute prediction from LiDAR observations at different timestamps. Although a shared sparse backbone allows these observations to interact in a common 3D feature space, their information is fused only implicitly through spatial convolution. It does not explicitly retrieve and aggregate cross-time features for individual Gaussian primitives. These two limitations motivate us to learn Gaussian structure adjustment while explicitly incorporating cross-time information into attribute prediction.

To address these limitations, we present Learning Gaussian Structure (LGS). For Gaussian structure adjustment, its \emph{Gaussian Densify Policy} predicts a Densify Map comprising a Prune Score and an Addition Score for each Gaussian primitive. During training, original, prune, and addition branches measure the local responses induced by the two operations and construct supervision for the policy. These intervention branches are discarded after training. For reliable Gaussian attribute prediction, we further introduce \emph{Cross-Time Point Query}, which retrieves neighboring features from Gaussian primitives at other timestamps and fuses them with the current decoder feature. The fused features guide Gaussian attribute prediction and are also provided to the Gaussian Densify Policy. Together, the two components learn the composition and attributes of the Gaussian representation.
Extensive experiments on Waymo and PandaSet demonstrate that LGS consistently outperforms existing methods in driving scene reconstruction. Component analyses verify the individual and complementary effects of Gaussian density adjustment and cross-time feature aggregation.

In summary, our main contributions are as follows:
\begin{itemize}
    \item We present LGS, which formulates feed-forward driving-scene reconstruction as learning both the discrete composition of the Gaussian set and the continuous attributes of each primitive.
    \item We propose a Gaussian Densify Policy that distills local responses induced by prune and add interventions into a Densify Map for Gaussian structure adjustment.
    \item We develop a Cross-Time Point Query that explicitly aggregates neighboring features from Gaussian primitives at other timestamps, providing complementary evidence for reliable Gaussian attribute prediction.
\end{itemize}

\section{Related Work}

\subsection{Feed-forward Gaussian reconstruction.}
Early driving-scene reconstruction methods rely on per-scene optimization and represent large-scale outdoor environments through static-dynamic decomposition, geometric priors, or specialized scene models~\citep{suds2023,unisim2023,neurad2023,emernerf2023,streetsurf2023,mars2023}. Although these methods achieve high-quality rendering, they require a separate optimization process for each scene. Feed-forward methods instead learn a generalizable mapping from sensor observations to scene representations. Early works predict pixel-aligned Gaussians from one or multiple images~\citep{pixelsplat2024,mvsplat2024,splatterimage2024}. Recent driving-scene methods construct 3D-aligned representations: EVolSplat~\citep{evolsplat2025} initializes volume-based primitives, UniSplat builds a latent 3D scaffold, PointForward~\citep{pointforward2026} places queries in world space, Ground4D~\citep{ground4d2026} adopts spatially grounded voxel queries, and Flux4D~\citep{flux4d2025} anchors Gaussians on LiDAR points aggregated across timestamps. Building on this LiDAR-anchored formulation, we explicitly enhance the composition of the initialized Gaussian set instead of only regressing its continuous states.

\subsection{Gaussian densification and pruning.}
The quality and efficiency of Gaussian Splatting depend on both primitive attributes and the composition of the primitive set. The original 3DGS~\citep{kerbl3dgs2023} adapts an initialized point cloud along a scene-specific optimization through gradient-based densification. Since feed-forward models lack such optimization, recent approaches transfer density control into learned reconstruction. One line of work learns where additional capacity should be generated. Generative Densification~\citep{nam2025generative} upsamples backbone features associated with selected coarse Gaussians and produces fine Gaussians in a single forward pass. F4Splat~\citep{kim2026f4splat} predicts per-region densification scores over multi-scale Gaussian maps and allocates a controllable Gaussian budget according to spatial complexity and multi-view overlap. Another line of work constructs the compact representation. SparseSplat~\citep{zhang2026sparsesplat} adapts entropy probabilistic sampling and predicts Gaussian attributes from local 3D neighborhoods. Complementarily, importance-based methods remove redundancy from an existing representation. RAP~\citep{yang2026rap} infers per-primitive importance from intrinsic Gaussian attributes and neighborhood statistics, enabling rendering-free feed-forward pruning.

These approaches transfer density control through generation, adaptive initialization, or importance-based removal. LGS instead treats LiDAR-derived Gaussians as a mutable candidate set and learns scene-dependent structural adjustments. Rather than using image entropy, a direct densification cue, or a generic importance objective, it supervises candidate-level interventions through changes in neighboring gradient responses. The learned policy transfers these decisions to unseen clips without intervention rendering or per-scene optimization at inference.


\begin{figure*}
  \centering
   \includegraphics[width=0.85\linewidth]{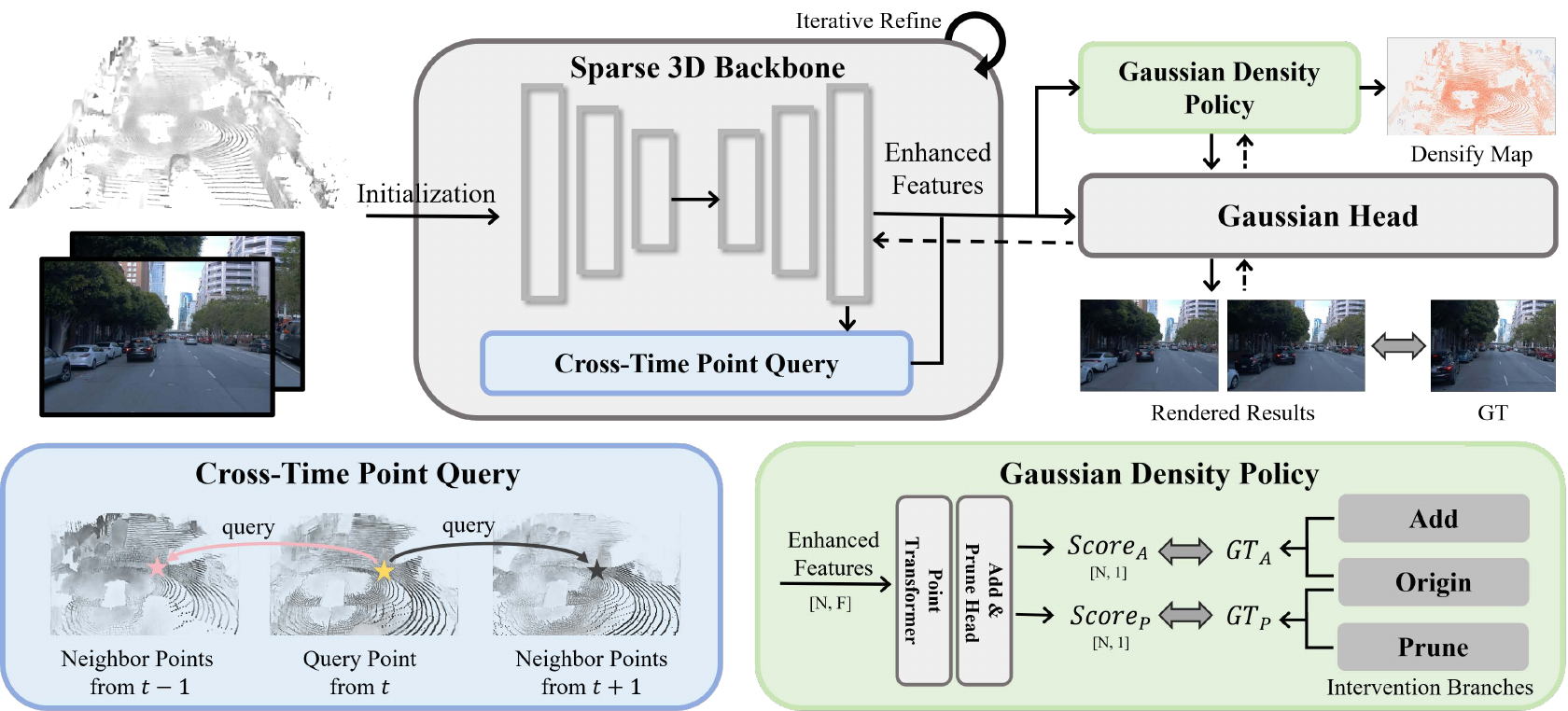}
   \caption{Overview of LGS. Given multi-camera images and LiDAR points
   captured at different timestamps, the
   sparse 3D backbone and Cross-Time Point Query produce enhanced point
   features, from which the Gaussian head iteratively refines primitive states.
   The Gaussian Densify Policy transforms these features into a Densify Map
   composed of Addition Scores and Prune Scores. The intervention branches
   supervise the two score components during training and are discarded at
   inference.}
   \label{fig:framework}
\end{figure*}

\section{Method}

\subsection{Overview}

\noindent\textbf{Preliminaries.}
The overall architecture follows the feed-forward Gaussian reconstruction frameworks of EVolSplat and Flux4D~\citep{evolsplat2025,flux4d2025}. Given a clip containing LiDAR points from $T$ timestamps and synchronized images from $C$ cameras, we aim to predict a dynamic Gaussian representation of the observed scene. Using the known sensor poses, we transform the LiDAR points captured at different timestamps into a shared coordinate system.
Specifically, gaussian primitive $i$ retains its acquisition timestamp $t_i$ and have the  following attributes:
\begin{equation}
g_i=[\mathbf{x}_i,\mathbf{q}_i,\mathbf{s}_i,\alpha_i,\mathbf{c}_i],
\end{equation}
where $\mathbf{x}$, $\mathbf{q}$, $\mathbf{s}$, $\alpha$, and $\mathbf{c}$ denote the position, rotation, scale, opacity, and color, respectively. A sparse-convolution backbone~\citep{torchsparse2022} takes the Gaussian attributes and timestamps as input and predicts an attribute residual $\Delta g_i$ together with an initial velocity $\mathbf{v}_i$. To render the scene at time $t$, each Gaussian is translated from its acquisition time $t_i$ according to the predicted velocity. The rendering loss is defined as:
\begin{equation}
\mathcal{L}_{\mathrm{render}}=
\lambda_{1}\mathcal{L}_{1}+
\lambda_{\mathrm{ssim}}\mathcal{L}_{\mathrm{SSIM}}+
\lambda_{d}\mathcal{L}_{\mathrm{depth}}+
\lambda_{v}\mathcal{L}_{\mathrm{velocity}}.
\end{equation}
Then, the backbone further updates both the Gaussian states and velocities for three iterations according to the rendering loss. Static and dynamic regions are therefore learned through the predicted motion rather than specified by an input partition. They additionally model the image background with a separate static Gaussian branch.

\noindent\textbf{Learning Gaussian Structure.}
Although this backbone predicts complete Gaussian attributes and motion, the number and membership of the Gaussian set remain fixed by initialization, while observations from other timestamps interact only through generic convolution. Building on this formulation, LGS learns both the composition of Gaussian representation and explicit cross-time evidence for each primitive. As illustrated in Figure~\ref{fig:framework}, we introduce Gaussian Densify Policy to adjust the primitive set and Cross-Time Point Query to retrieve informative features from other timestamps.

\subsection{Gaussian Densify Policy}
\label{sec:density}

\begin{figure}
  \centering
   \includegraphics[width=1.\linewidth]{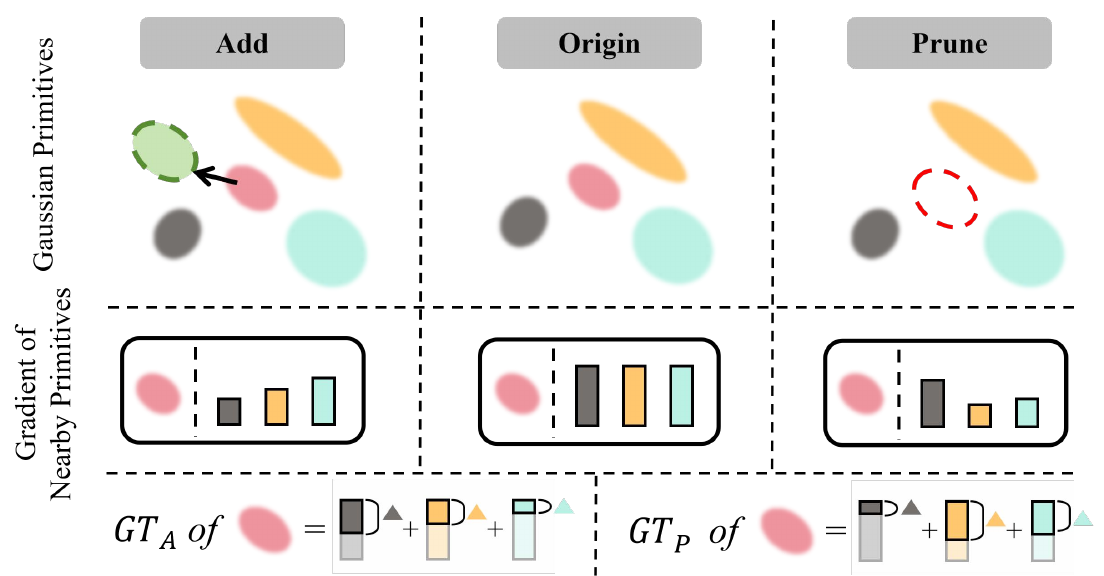}
    \caption{Intervention supervision for the Gaussian Densify Policy. For each
    sampled candidate, shown in red, the Add and Prune branches modify the
    primitive set relative to the Origin branch. The resulting changes in the
    gradients of nearby primitives are aggregated to construct the Addition
    target $GT_A$ and Prune target $GT_P$.}
   \label{fig:density}
\end{figure}

The rendering loss provides gradients for updating the attributes of existing Gaussian primitives, but it does not directly indicate whether a primitive should be pruned or a new primitive should be added. Moreover, a pruned primitive is removed from the representation and can no longer receive gradients. To obtain direct supervision for these operations, we apply controlled prune and add interventions and measure their effects on the rendering gradients of neighboring primitives. The resulting responses are used to supervise the Gaussian Densify Policy, as illustrated in Figure~\ref{fig:density}.

\noindent\textbf{Gaussian interventions.}
At each training iteration, we sample a subset of primitives with probability $\rho$ and repeat this process for several rounds to improve coverage. Starting from the same predicted Gaussian set, we construct three branches. The original branch remains unchanged. The prune branch prunes the sampled primitives, while the addition branch adds a perturbed copy of each sampled primitive. All three branches use the same target views and rendering objective. The representation network is frozen, so the prune and add operations are the only differences among the branches. These branches are used only to construct the training targets.

\noindent\textbf{Local gradient response.}
Let $\mathcal{B}_1$, $\mathcal{B}_2$, and $\mathcal{B}_3$ denote the original, prune, and addition branches, respectively. For a Gaussian primitive $i$ selected for intervention, $j$ indexes its neighboring primitives and $l$ indexes all Gaussian primitives in branch $\mathcal{B}$. Their Gaussian states are denoted by $g_i$, $g_j$, and $g_l$, respectively. We denote the rendering loss $\mathcal{L}_{render}$ evaluated for branch $\mathcal{B}$ and compute its gradients with respect to the Gaussian attributes. Since the gradients associated with different attributes may vary considerably in scale, we normalize the gradient of every attribute by its maximum absolute value across all Gaussian primitives. We then average the normalized gradient magnitudes to obtain a scalar response:
\begin{equation}
r_j^{\mathcal{B}}=
\frac{1}{D}\sum_{c=1}^{D}
\left|
\frac{(\nabla_{g_j}\mathcal{L}^{B})_c}
{\max_l|(\nabla_{g_l}\mathcal{L}^{B})_c|+\epsilon}
\right|,
\end{equation}
where $D$ denotes the number of Gaussian attribute dimensions and $\epsilon$ ensures numerical stability. For the prune operation on Gaussian primitive $i$, we find its $K$ spatial neighbors that remain in the prune branch and compute:
\begin{equation}
\Delta_i^{\mathrm{prune}}=
\sum_{j\in K_{(i)}}
\left(r_j^{\mathcal{B}_2}-r_j^{\mathcal{B}_1}\right).
\end{equation}
For the add operation, we measure the response change of the original neighboring primitives after a perturbed copy of primitive $i$ is introduced in the addition branch:
\begin{equation}
\Delta_i^{\mathrm{add}}=
\sum_{j\in K_{(i)}}
\left(r_j^{\mathcal{B}_3}-r_j^{\mathcal{B}_1}\right).
\end{equation}
The neighboring Gaussian primitives provide a consistent reference between the original and intervention branches. A negative $\Delta_i$ indicates that operation reduces the gradient response within the local neighborhood, while a positive value indicates the opposite effect. We apply a signed logarithmic transform followed by z-score normalization and sigmoid mapping to obtain a supervision target $y_i^a\in[0,1]$, where $a\in\{\mathrm{prune},\mathrm{add}\}$. A larger $y_i^a$ indicates that the corresponding intervention is more beneficial.

\noindent\textbf{Policy prediction and optimization.}
To predict the intervention targets in unseen scenes, we implement the densify policy with Point Transformer V3~\citep{ptv3_2024}. Its input concatenates decoder feature $\mathbf f'_i$ with normalized position, Gaussian attributes, and voxel-density feature $d_i$. A shared encoder propagates spatial context among Gaussian primitives, and two prediction heads estimate the Prune and Addition Scores that form the Densify Map:
\begin{equation}
(p_i^{\mathrm{prune}},p_i^{\mathrm{add}})=
\sigma\left(D(\mathbf f'_i,g_i,d_i)\right),
\end{equation}
where $D$ denotes the shared spatial encoder and two prediction heads. 
We optimize the two scores over Gaussian primitives with valid targets using a confidence-weighted regression loss:
\begin{equation}
\begin{aligned}
\mathcal{L}_{\mathrm{density}}&=
\sum_{a\in\{\mathrm{prune},\mathrm{add}\}}
\frac{\sum_i w_i(p_i-y_i)^2}
{\sum_i w_i+\epsilon},\\
w_i&=(2|y_i-0.5|)^\gamma.
\end{aligned}
\end{equation}
Given that $y_i\in[0,1]$, the weight $w_i$ suppresses ambiguous targets near $0.5$ and emphasizes confident intervention responses, while $\gamma$ controls the weighting strength.

\noindent\textbf{Feed-forward density adjustment.}
During inference, the intervention branches are discarded, and the Gaussian Densify Policy predicts the Prune and Addition Scores in a single forward pass. Given the thresholds $\theta_p$ and $\theta_a$, we update the Gaussian set as:
\begin{equation}
\begin{aligned}
\mathcal{G}'={}&
\{g_i\mid p_i^{\mathrm{prune}}\leq\theta_p\}\\
&{}\cup
\{\operatorname{Add}(g_i)\mid
p_i^{\mathrm{add}}>\theta_a,\,
p_i^{\mathrm{prune}}\leq\theta_p\},
\end{aligned}
\end{equation}
where $\operatorname{Add}(g_i)$ denotes a perturbed copy of Gaussian primitive $i$. The first term contains the primitives retained after the prune operation, while the second adds new primitives selected by the add operation. When both scores exceed their thresholds, prune takes precedence over add. Based on the ablation study in Table~\ref{tab:density_budget}, we set both $\theta_p$ and $\theta_a$ to $0.7$ to balance reconstruction quality and the number of Gaussian primitives. This procedure adjusts the composition of the Gaussian representation in a feed-forward manner without intervention rendering at inference.

\subsection{Cross-Time Point Query}
\label{sec:temporal}

The shared sparse backbone allows spatially adjacent Gaussian primitives to exchange information, but this implicit interaction does not ensure that each primitive exploits evidence from another timestamp. To address this limitation, we introduce Cross-Time Point Query, which explicitly retrieves local features from other timestamps.

\begin{figure*}[t]
  \centering
   \includegraphics[width=1.\linewidth]{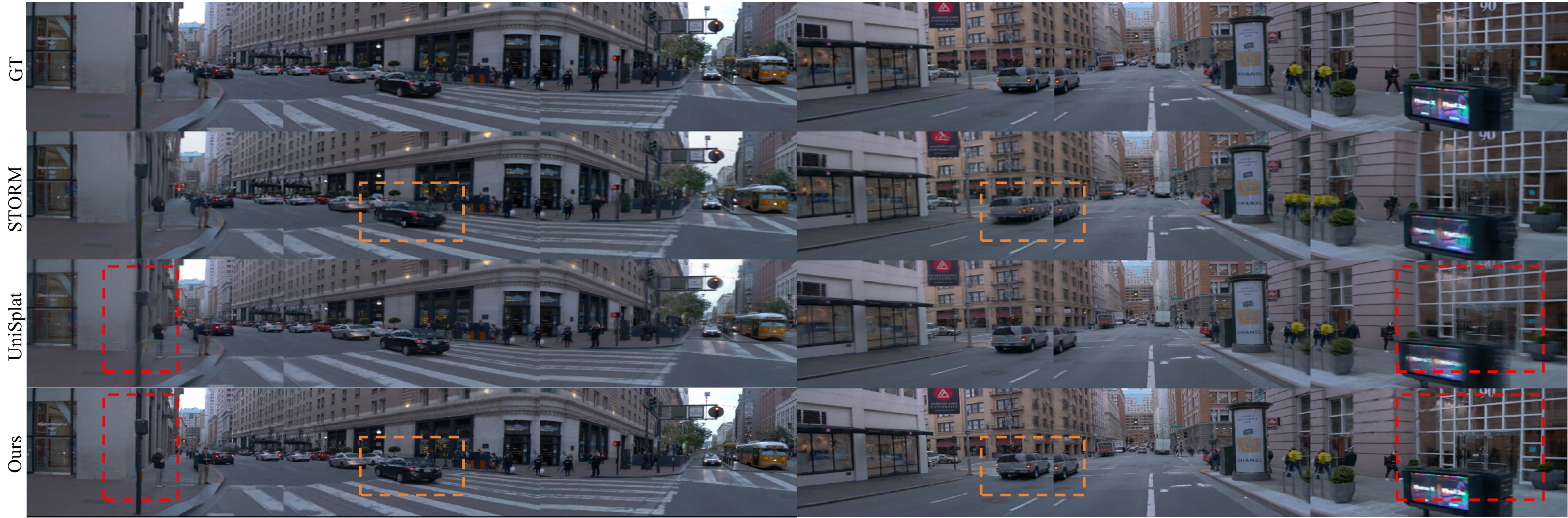}
   \caption{Qualitative comparison of novel-view synthesis on Waymo. Compared
   with STORM~\citep{storm2025} and UniSplat~\citep{unisplat2025}, LGS
   reconstructs moving vehicles in the orange boxes with sharper boundaries
   and preserves finer static structures in the red boxes, reducing blur and
   structural artifacts.}
   \label{fig:waymo_qualitative}
\end{figure*}

\noindent\textbf{Cross-time feature aggregation.}
Let $\mathbf{f}_i$, $\mathbf{x}_i$, and $t_i$ denote the decoder feature, position, and timestamp of Gaussian primitive $i$, respectively. For each primitive, we retrieve a local neighborhood from other timestamps. Specifically, $\mathcal{N}(i)^{t}$ contains the $K$ nearest Gaussian primitives to primitive $i$ whose timestamps satisfy $t_j\neq t_i$.

We apply mean pooling to the retrieved features and fuse the resulting cross-time feature with the original decoder feature through a residual projection:
\begin{equation}
\bar{\mathbf f}_{i}=
\frac{1}{|\mathcal{N}(i)^{t}|}
\sum_{j\in\mathcal{N}(i)^{t}}\mathbf f_{j},
\qquad
\mathbf f'_i=\mathbf f_i+W_{proj}[\mathbf f_i,\bar{\mathbf f}_{i}],
\end{equation}
where $\bar{\mathbf f}_{i}$ denotes the pooled cross-time feature, $[\cdot,\cdot]$ denotes feature concatenation, and $W_{proj}$ is the fusion projection. Mean pooling summarizes the retrieved neighborhood regardless of primitive order. We concatenate the resulting cross-time context with the decoder feature and apply a residual projection, retaining the original representation while incorporating information from other timestamps. The fused feature $\mathbf f'_i$ is then used to predict the Gaussian residual, allowing cross-time evidence to guide Gaussian state prediction, and is also provided as input to the Gaussian Densify Policy.

\begin{table}[t]
\centering
\resizebox{\columnwidth}{!}{
\begin{tabular}{lccccccc}
\toprule
& \multicolumn{3}{c}{Full Image} & \multicolumn{2}{c}{Dynamic} & \multicolumn{2}{c}{Static} \\
\cmidrule(lr){2-4}\cmidrule(lr){5-6}\cmidrule(lr){7-8}
Method & PSNR$\uparrow$ & SSIM$\uparrow$ & LPIPS$\downarrow$ & PSNR$\uparrow$ & SSIM$\uparrow$ & PSNR$\uparrow$ & SSIM$\uparrow$ \\
\midrule
LGM~\citep{lgm2024} & 17.49 & 0.47 & 0.33 & 17.79 & 0.49 & 15.37 & 0.39 \\
PixelSplat~\citep{pixelsplat2024} & 18.24 & 0.56 & 0.30 & 18.63 & 0.58 & 16.96 & 0.44 \\
MVSplat~\citep{mvsplat2024} & 19.00 & 0.57 & 0.28 & 19.29 & 0.58 & 17.35 & 0.47 \\
L4GM~\citep{l4gm2024} & 17.63 & 0.54 & 0.31 & 18.58 & 0.56 & 16.78 & 0.43 \\
DrivingRecon~\citep{drivingrecon2024} & 20.63 & 0.61 & 0.21 & 20.97 & 0.62 & 19.70 & 0.51 \\
GaussianSTORM~\citep{storm2025} & 25.40 & 0.779 & 0.189 & 22.82 & 0.635 & 25.67 & 0.785 \\
UniSplat~\citep{unisplat2025} & 26.28 & 0.818 & 0.150 & 24.37 & 0.711 & 26.83 & 0.825 \\
\midrule
LGS$_{base}$ (Ours) & 23.40 & 0.712 & 0.207 & 21.07 & 0.614 & 23.64 & 0.743 \\
LGS (Ours) & \textbf{28.04} & \textbf{0.885} & \textbf{0.113} & \textbf{26.54} & \textbf{0.848} & \textbf{28.22} & \textbf{0.910} \\
\bottomrule
\end{tabular}
}
\caption{Quantitative comparison on the Waymo Dataset. LGS achieves the best rendering quality over the full image and both dynamic and static regions. Bests are in highlight.}
\label{tab:main}
\end{table}

\subsection{Joint Training and Feed-Forward Inference}
\label{sec:joint_training}

Jointly optimizing the backbone network, Cross-Time Point Query, and Gaussian Densify Policy from scratch can destabilize policy supervision as the intervention targets change with the predicted Gaussian states. We therefore adopt a three-stage training strategy. First, we train the backbone and Cross-Time Point Query using rendering supervision, with optional density perturbations to improve robustness to different Gaussian layouts. Second, we freeze these modules and optimize the Gaussian Densify Policy module with the intervention targets. Finally, we freeze the policy module and fine-tune the backbone with the predicted prune and add operations. This strategy separates target construction from representation learning.
The intervention branches are used only to construct policy targets and are discarded after training. During inference, the backbone and Cross-Time Point Query predict the Gaussian states, after which the Gaussian Densify Policy produces the Densify Map containing the Prune and Addition Scores. Based on these scores, we apply the predicted prune and add operations before final refinement and rendering, without intervention rendering.

\section{Experiments}

\subsection{Experimental Protocol}

\noindent\textbf{Waymo protocol.}
We conduct experiments on the Waymo Open Dataset~\citep{waymo2020} following the scene split and novel-view synthesis protocol of DrivingRecon~\citep{drivingrecon2024}. We use three front-facing cameras at a resolution of $256\times512$, and three-frame input clips. Every tenth frame is held out from the input and rendered from its neighboring context frames. We report PSNR, SSIM, and LPIPS over the full image, together with PSNR and SSIM over dynamic and static regions.

\noindent\textbf{PandaSet protocol.}
We additionally train and evaluate LGS on PandaSet~\citep{pandaset2021} following the scene split and evaluation protocol of Flux4D~\citep{flux4d2025}. We use full-resolution images from the front camera. Each clip takes six frames as input, whose LiDAR observations are aggregated to initialize the Gaussian representation, while the held-out target frames provide rendering supervision during training and are used for evaluation. We report PSNR, SSIM, and depth mean absolute error (DMAE) over dynamic regions and full images.


\noindent\textbf{Implementation details.}
The sparse convolution backbone uses a voxel size of $0.1$ m, and Gaussian refinement is performed for three iterations. We set $(\lambda_1,\lambda_{\mathrm{ssim}},\lambda_d,\lambda_v)$ to $(0.8,0.2,0.01,0.01)$. At each of the three intervention rounds, we sample $20\%$ of the Gaussian primitives initialized from LiDAR points. For each sampled primitive, we use 8 spatial neighbors to construct the intervention targets, while Cross-Time Point Query retrieves 8 neighboring primitives from other timestamps for feature aggregation. The add operation perturbs each copied position with zero-mean Gaussian noise of standard deviation $0.01$ m. We set $\gamma=2$ for confidence weighting. The Gaussian Densify Policy uses a Point Transformer V3 patch size of $128$, followed by four hidden blocks of width $32$ in each prediction head. Following the three-stage strategy in Section~\ref{sec:joint_training}, the three stages are trained for $20{,}000$, $15{,}000$, and $20{,}000$ iterations, respectively, using 8 GPUs. All variants share the same outdoor background initialization based on LiDAR-aligned monocular depth~\citep{depthanything3} and sampled sky points.

\subsection{Main Results}


\begin{table}[t]
\centering
\resizebox{\columnwidth}{!}{
\begin{tabular}{lccccccc}
\toprule
& \multicolumn{3}{c}{Dynamic Only} & \multicolumn{3}{c}{Full Image} & \multicolumn{1}{c}{Recon. Speed} \\
\cmidrule(lr){2-4}\cmidrule(lr){5-7}\cmidrule(lr){8-8}
Method & PSNR$\uparrow$ & SSIM$\uparrow$ & DMAE$\downarrow$
& PSNR$\uparrow$ & SSIM$\uparrow$ & DMAE$\downarrow$ & Time$\downarrow$ \\
\midrule
EmerNeRF~\citep{emernerf2023} & 17.79 & 0.411 & 6.09 &  22.80 & 0.624 & 4.24 & ~100min \\
DeSiRe-GS~\citep{desiregs2025} & 19.08 &  0.477 & 3.36 &   22.25 & 0.608 & 24.89 &  ~120min \\
DepthSplat~\citep{depthsplat2025} & 16.87 & 0.425 & 6.18 & 21.40 & 0.595 & 2.73 & 0.87s \\
L4GM~\citep{l4gm2024} & 17.36 & 0.343 & -- & 19.38 & 0.465 & -- & 0.32s \\
STORM~\citep{storm2025} & 17.65 & 0.367 & 5.24 & 20.79 & 0.508 & 4.80 & \textbf{0.08s} \\
\midrule
LGS$_{base}$(Ours) & 19.56 & 0.500 & 3.95 & 21.23 & 0.644 & 2.35 & 0.37s \\
LGS (Ours) & \textbf{22.43} & \textbf{0.678} & \textbf{2.06} & \textbf{25.03} & \textbf{0.762} & \textbf{1.52} & 1.92s \\
\bottomrule
\end{tabular}}
\caption{Quantitative comparison on PandaSet~\citep{pandaset2021} dataset. LGS achieves the best reconstruction quality over dynamic regions and full images. Bests are in highlight.}
\label{tab:pandaset}
\end{table}

\begin{figure}
  \centering
   \includegraphics[width=1.\linewidth]{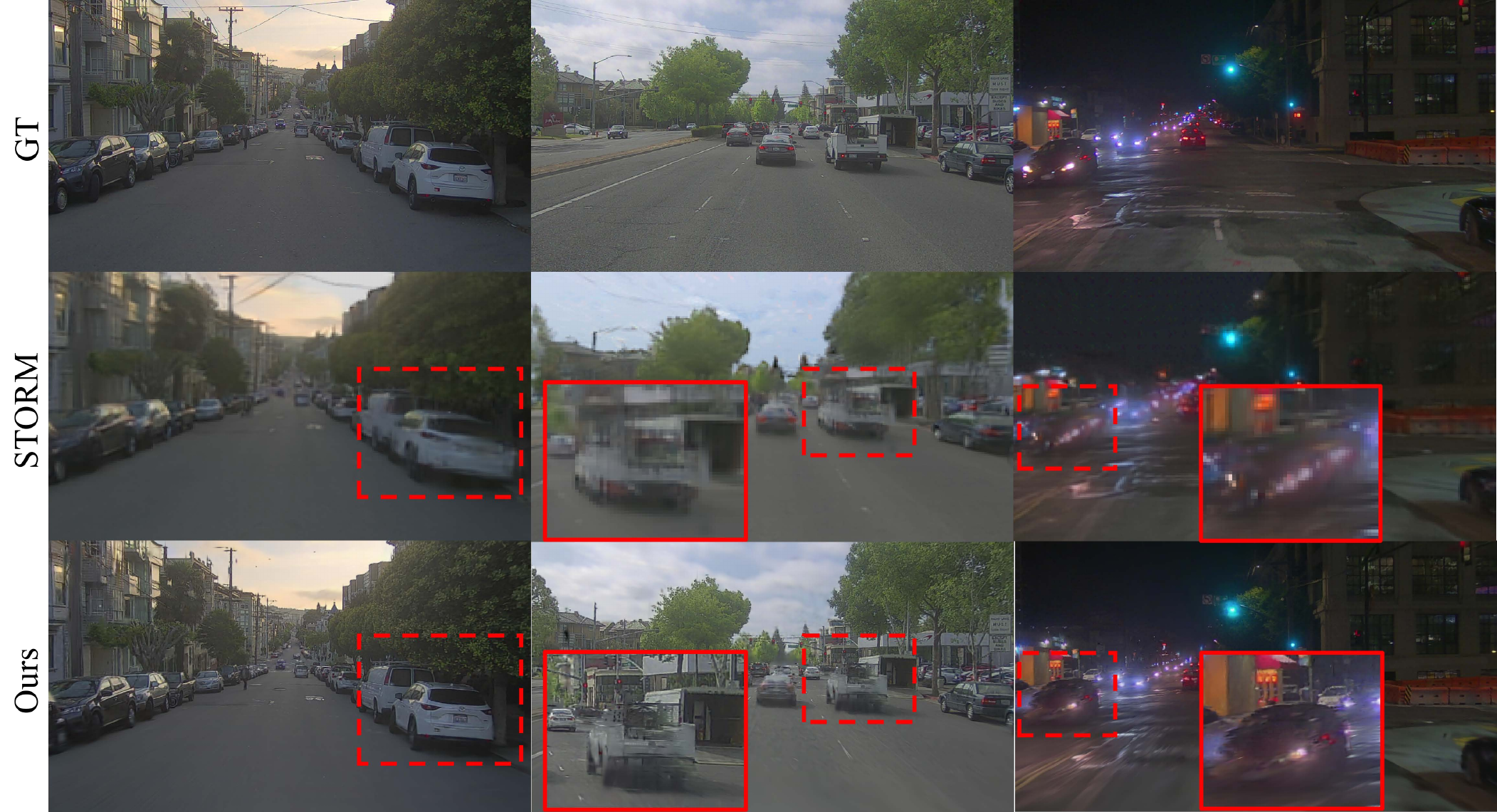}
    \caption{Qualitative comparison on PandaSet. Compared with
    STORM~\citep{storm2025}, LGS better preserves vehicle shapes and fine scene
    details while reducing blur in the highlighted regions. Red boxes mark
    challenging areas and their enlarged views.}
   \label{fig:pandaset_qualitative}
\end{figure}

\noindent\textbf{Waymo Dataset.}
As reported in Table~\ref{tab:main}, LGS achieves the best results across all evaluation metrics on Waymo. Compared with UniSplat, LGS improves full-image PSNR from 26.28\,dB to 28.04\,dB and reduces LPIPS from 0.150 to 0.113. The improvements over both dynamic and static regions show that LGS benefits moving objects as well as the surrounding scene. As shown in Figure~\ref{fig:waymo_qualitative}, LGS reconstructs moving vehicles with sharper boundaries and preserves finer structures.

\noindent\textbf{PandaSet Dataset.}
On PandaSet, LGS also achieves the best results across all six quality metrics, as shown in Table~\ref{tab:pandaset}. In particular, it improves dynamic-region PSNR from 17.65\,dB to 22.43\,dB and full-image PSNR from 21.40\,dB to 25.03\,dB, while substantially reducing the corresponding depth errors. Figure~\ref{fig:pandaset_qualitative} shows that these improvements translate into better preserved vehicle shapes and finer scene details. LGS$_{base}$ denotes the representation before iterative refinement. The final refinement further improves reconstruction quality, with an increase in reconstruction time from 0.37\,s to 1.92\,s.

\subsection{Component Analysis}
To evaluate the contribution of the Gaussian Densify Policy and Cross-Time Point Query, we conduct the $2\times2$ factorial ablation shown in Table~\ref{tab:factorial}.

\begin{table}[t]
\centering
\resizebox{\columnwidth}{!}{
\begin{tabular}{cc|ccc|ccc}
\toprule
GDP & CTPQ & PSNR$\uparrow$ & SSIM$\uparrow$ & LPIPS$\downarrow$
& Gaussians (K) & Params (M) & Time (s/frame) \\
\midrule
& & 26.55 & 0.857 & 0.138 & 519.8 & 86.96 & 1.77 \\
\checkmark & & 27.68 & 0.878 & 0.119 & 584.2 & 95.65 & 1.82 \\
& \checkmark & 27.31 & 0.862 & 0.130 & 519.8 & 86.98 & 1.87 \\
\checkmark & \checkmark & \textbf{28.04} & \textbf{0.885} & \textbf{0.113} & 584.2 & 95.67 & 1.90 \\
\bottomrule
\end{tabular}}
\caption{Ablation and efficiency analysis of the Gaussian Densify Policy (GDP) and Cross-Time Point Query (CTPQ).}
\label{tab:factorial}
\end{table}

\begin{figure}[t]
  \centering
   \includegraphics[width=1.\linewidth]{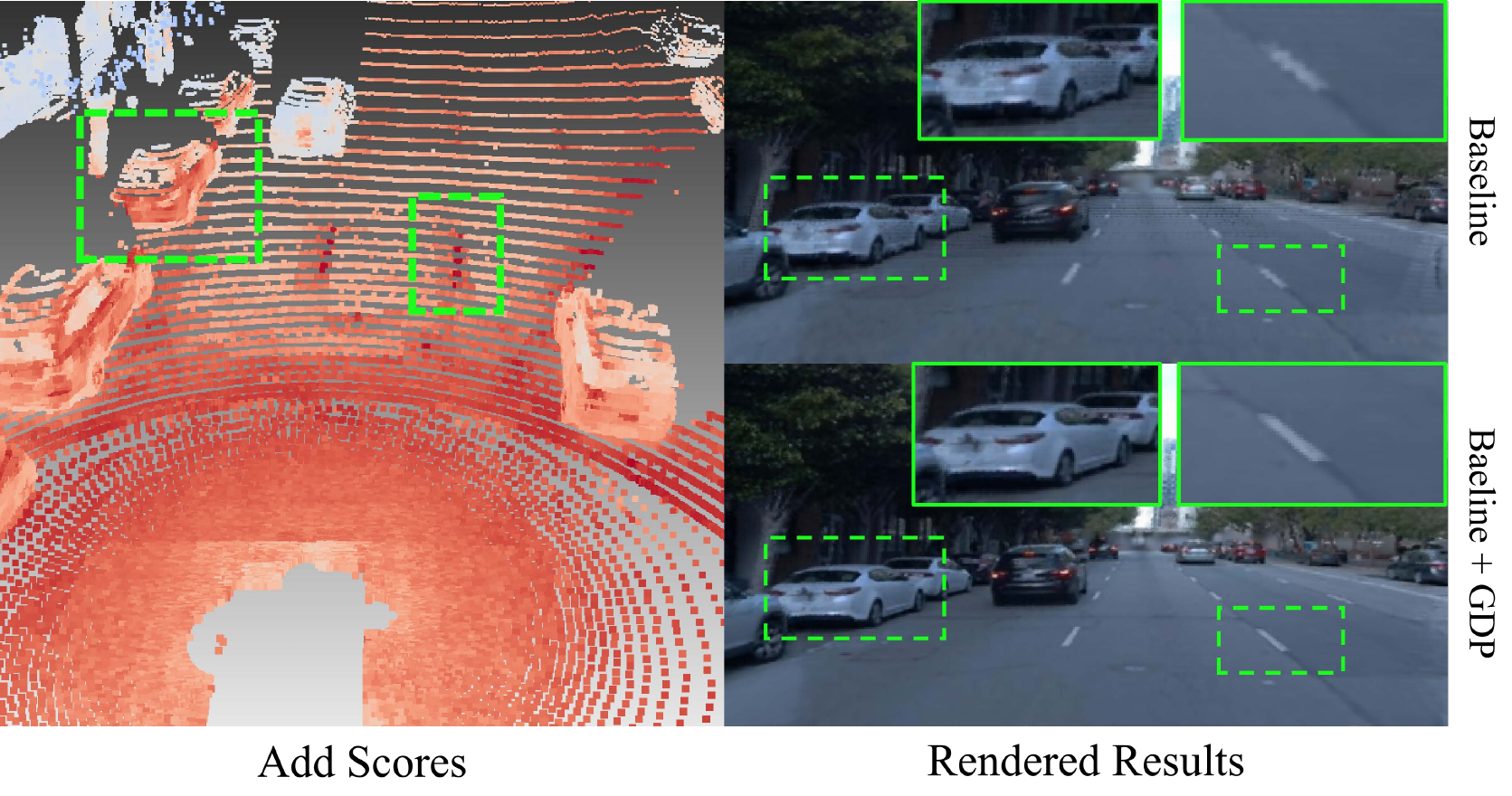}
    \caption{Effect of the Addition Scores predicted by the Gaussian Densify
    Policy. The scores locate regions with rich texture variations, including
    vehicle surfaces and lane markings highlighted by the green boxes. The
    resulting density adjustment improves the reconstruction of fine details
    in the rendered results.}
   \label{fig:gdp_visualization}
\end{figure}

Adding the Gaussian Densify Policy improves PSNR from 26.55\,dB to 27.68\,dB and reduces LPIPS from 0.138 to 0.119. Meanwhile, the learned prune and add decisions adjust the Gaussian count from 519.8K to 584.2K, showing the benefit of restructuring the Gaussian representation. Cross-Time Point Query improves PSNR to 27.31\,dB without changing the number of Gaussian primitives, indicating that explicitly retrieved cross-time evidence improves Gaussian state prediction rather than representation capacity. Combining both components achieves the best performance, with 28.04\,dB PSNR, 0.885 SSIM, and 0.113 LPIPS, which verifies their complementary effects. GDP introduces 8.69M parameters and 0.05\,s additional inference time per frame, while CTPQ adds only 0.02M parameters and 0.10\,s per frame. The full model contains 95.67M parameters and runs at 1.90\,s.

\subsection{Gaussian Densify Policy Analysis}

\begin{figure}[t]
  \centering
   \includegraphics[width=1.\linewidth]{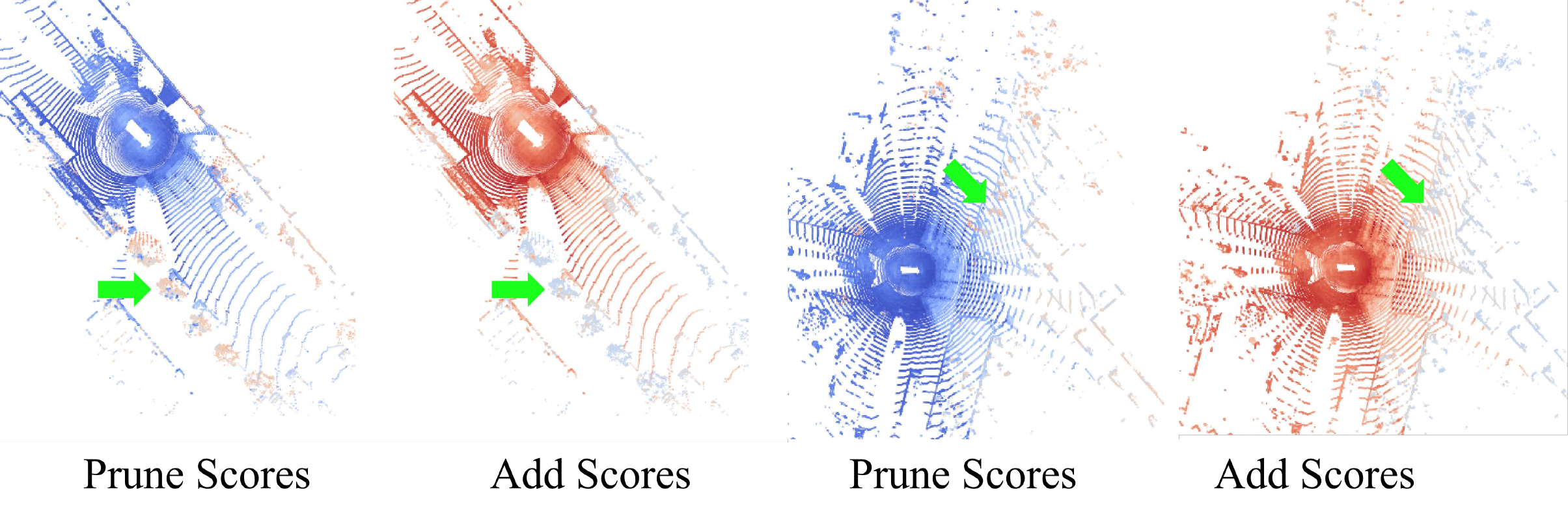}
    \caption{Relation between the Addition and Prune Scores.
    Although their supervision targets are constructed separately, the two
    outputs exhibit spatially complementary responses. As highlighted by the
    green arrows, regions assigned high Addition Scores generally receive low
    Prune Scores, and vice versa, distinguishing Gaussian primitives that
    benefit from additional representation capacity from those more suitable
    for removal. Better visualized with zoom in.}
   \label{fig:densify_scores_visualization}
\end{figure}

\noindent\textbf{Densify Map visualization.}
We first visualize the scores predicted by the Gaussian Densify Policy. As shown in Figure~\ref{fig:gdp_visualization}, the Addition Scores respond strongly to vehicle surfaces and lane markings. The corresponding rendered results recover clearer vehicle boundaries and lane details, indicating that the addition branch identifies regions that benefit from additional representation capacity. Figure~\ref{fig:densify_scores_visualization} further compares the Addition and Prune Scores. Although their supervision targets are constructed independently, the two outputs exhibit spatially complementary responses without an explicit constraint between them. Regions with high Addition Scores generally receive low Prune Scores, and vice versa. This behavior enables the policy to distinguish regions requiring additional capacity from primitives that are suitable for removal.

\begin{table}[t]
\centering
\resizebox{\columnwidth}{!}{
\begin{tabular}{ccccc|ccc}
\toprule
$\theta_p$ & $\theta_a$ & Pruned (\%) & Added (\%) &
Gaussians (K) & PSNR$\uparrow$ & SSIM$\uparrow$ & LPIPS$\downarrow$ \\
\midrule
\multicolumn{2}{c}{Baseline} & OFF & OFF & 519.8 & 27.31 & 0.862 & 0.130 \\
\midrule
OFF & 0.50 & 0.00 & 99.78 & 1038.5 & 27.83 & 0.871 & 0.120 \\
OFF & 0.60 & 0.00 & 83.61 & 954.4 & 28.03 & 0.885 & 0.114 \\
OFF & 0.70 & 0.00 & 14.91 & 597.3 & 28.09 & 0.888 & 0.110 \\
OFF & 0.80 & 0.00 & 6.21  & 552.1  & 27.72 & 0.873 & 0.123 \\
\midrule
0.50 & OFF & 7.71 & 0.00 & 479.7 & 26.94 & 0.846 & 0.144 \\
0.60 & OFF & 4.34 & 0.00 & 497.2 & 27.20 & 0.857 & 0.135 \\
0.70 & OFF & 2.25 & 0.00 & 508.1 & 27.29 & 0.861 & 0.131 \\
0.80 & OFF & 0.31 & 0.00 & 518.2 & 27.31 & 0.862 & 0.131 \\
\midrule
\multicolumn{2}{c}{Random} & 2.25 & 14.91 & 584.2 & 27.52 & 0.869 & 0.125 \\
0.70 & 0.70 & 2.25 & 14.91 & 584.2 & 28.04 & 0.885 & 0.113 \\
\bottomrule
\end{tabular}}
\caption{Analysis of pruning and addition thresholds. Random decisions use matched action rates and a comparable Gaussian count. We set both thresholds to 0.7 to balance reconstruction quality and representation size. }
\label{tab:density_budget}
\end{table}

\noindent\textbf{Density thresholds and Gaussian budget.}
We analyze the pruning and addition thresholds in Table~\ref{tab:density_budget}. Addition achieves the best reconstruction quality at $\theta_a=0.7$, while lower thresholds substantially increase the Gaussian count without further improvement. The prune branch behaves more conservatively. Setting $\theta_p=0.7$ removes $2.25\%$ of the primitives with only a minor change in reconstruction quality. Notably, random actions improve PSNR from 27.31\,dB to 27.52\,dB over the baseline by increasing the Gaussian count from 519.8K to 584.8K. This result suggests that the fixed initialized Gaussian set provides insufficient representation capacity in some regions. With the same action rates and a comparable Gaussian count, the learned policy further improves PSNR to 28.04\,dB
, showing that the selected locations are more important than representation size alone. Compared with addition alone at $\theta_a=0.7$, the joint setting uses 13.1K fewer Gaussians with only a 0.05\,dB decrease in PSNR. We therefore set both thresholds to 0.7 to balance reconstruction quality and representation size.

\subsection{Cross-Time Point Query Analysis}

\begin{figure}[t]
  \centering
   \includegraphics[width=1.\linewidth]{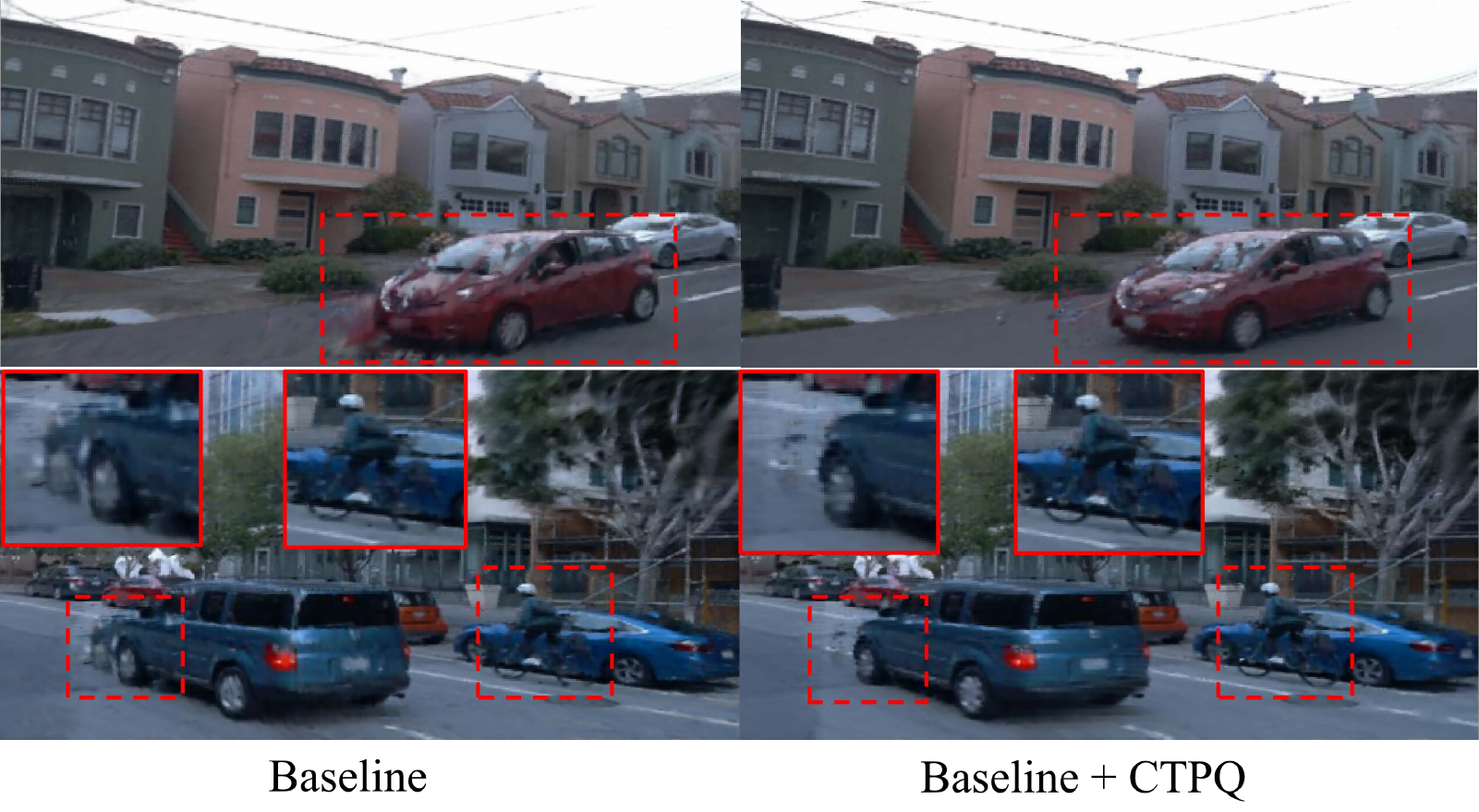}
    \caption{Effect of Cross-Time Point Query. Compared with the baseline on
    the left, CTPQ retrieves neighboring evidence from other timestamps and
    improves the reconstruction of moving vehicles and riders on the right.
    Red boxes highlight the preserved object shapes and reduced blur.}
   \label{fig:ctpq_visualization}
\end{figure}

As shown in Table~\ref{tab:factorial}, introducing Cross-Time Point Query improves PSNR from 26.55\,dB to 27.31\,dB and reduces LPIPS from 0.138 to 0.130 without changing the number of Gaussian primitives. When combined with the Gaussian Densify Policy, CTPQ further improves PSNR from 27.68\,dB to 28.04\,dB and LPIPS from 0.119 to 0.113. These consistent gains show that explicit cross-time retrieval complements both the shared spatial backbone and Gaussian density adjustment. As shown in Figure~\ref{fig:ctpq_visualization}, CTPQ better preserves the shapes of moving vehicles and riders while reducing blur in the highlighted regions.

\section{Discussion and Limitations}

Although LGS improves Gaussian set construction and state prediction, several limitations remain. The fixed density thresholds may not provide the optimal balance for every scene, while background construction still relies on LiDAR-aligned monocular depth and sampled sky points. Moreover, Euclidean cross-time retrieval may mix different physical structures for fast-moving objects and introduces additional search cost. Future work will explore scene-dependent density adjustment, unified learnable background modeling, and more efficient motion-aware retrieval.

\section{Conclusion}

In this paper, we presented LGS, a feed-forward framework that learns both the composition of Gaussian representations and primitive attributes for driving scene reconstruction. Its Gaussian Densify Policy learns structural adjustment from local rendering gradient responses induced by prune and add interventions, while Cross-Time Point Query explicitly aggregates neighboring features from Gaussian primitives at other timestamps for reliable attribute prediction. Experiments on Waymo and PandaSet demonstrate consistent improvements in reconstruction quality over existing methods for novel view synthesis, and component analyses verify the effectiveness and complementary roles of both designs.

\bibliography{aaai2027}
\end{document}